\documentclass[sn-basic, iicol]{sn-jnl}

\jyear{2023}%

\theoremstyle{thmstyleone}%
\theoremstyle{thmstyletwo}%

\theoremstyle{thmstylethree}%

\usepackage{siunitx} %
\usepackage{hyperref}
\usepackage{cleveref} %
\usepackage{tabularray} %
\usepackage{makecell} %
\usepackage{pifont}%
\usepackage{siunitx} %
\usepackage{graphicx}
\usepackage{mathtools} %

\newcommand{\cmark}{\ding{51}} %

\renewcommand\vec{\mathbf}
\newcommand{\fB}{\mathcal{B}}
\newcommand{\fO}{\mathcal{O}}
\newcommand{\fI}{\mathcal{I}}
\newcommand{\fP}{\mathcal{P}}
\newcommand{\fT}{\mathcal{T}}
\newcommand{\fTs}{\mathcal{T}_\mathcal{S}}
\newcommand{\fOs}{\mathcal{O}_\mathcal{S}}
\newcommand{\Toi}{\vec{T}_{\fO\fI}}
\newcommand{\Tbo}{\vec{T}_{\fB\fO}}
\newcommand{\Tbt}{\vec{T}_{\fB\fT}}

\newcommand{\symbolA}[3]{{}_{#1}\boldsymbol{#2}_{#3}}
\newcommand{\symbolB}[2]{\boldsymbol{#1}_{#2}}

\newcommand{\Mtask}{\mathcal{X}}
\newcommand{\dtask}{d_\mathcal{X}}

\newcommand{\Mconf}{\mathcal{C}}

\usepackage[nolist,nohyperlinks]{acronym}
\begin{acronym}
\acro{MAV}{Micro Aerial Vehicle}
\newacroindefinite{MAV}{an}{a}
\acro{OMAV}{Omnidirectional Micro Aerial Vehicle}
\newacroindefinite{OMAV}{an}{an}
\acro{UAV}{Uncrewed Aerial Vehicle}
\acro{IMU}{Inertial Measurement Unit}
\newacroindefinite{IMU}{an}{an}
\acro{MAE}{Mean Absolute Error}
\acro{VIO}{Visual Inertial Odometry}
\acro{NDT}{Non-destructive testing}
\acro{RMP}{Riemannian Motion Policie} %

\end{acronym}

\begin{document}

\title[A Complete Aerial Layouting System]{Chasing Millimeters: Design, Navigation and State Estimation for Precise In-flight Marking on Ceilings}

\author*{\fnm{Christian} \sur{Lanegger}}\email{clanegge@ethz.ch}
\equalcont{These authors contributed equally to this work.}

\author*{\fnm{Michael} \sur{Pantic}}\email{mpantic@ethz.ch}
\equalcont{These authors contributed equally to this work.}

\author{\fnm{Rik} \sur{B\"{a}hnemann}}\email{brik@ethz.ch}

\author{\fnm{Roland} \sur{Siegwart}}\email{rsiegwart@ethz.ch}

\author{\fnm{Lionel} \sur{Ott}}\email{lioott@ethz.ch}

\affil{\orgdiv{Autonomous Systems Lab}, \orgname{ETH Zurich}, \orgaddress{\street{Leonhardstrasse 21}, \city{Zurich}, \postcode{8092}, \country{Switzerland}}}

\abstract{Precise markings for drilling and assembly are crucial, laborious construction tasks. Aerial robots with suitable end-effectors are capable of markings at the millimeter scale. However, so far, they have only been demonstrated under laboratory conditions where rigid state estimation and navigation assumptions do not impede robustness and accuracy. This paper presents a complete aerial layouting system capable of precise markings on-site under realistic conditions. We use a compliant actuated end-effector on an omnidirectional flying base. Combining a two-stage factor-graph state estimator with a Riemannian Motion Policy-based navigation stack, we avoid the need for a globally consistent estimate and increase robustness. The policy-based navigation is structured into individual behaviors in different state spaces. Through a comprehensive study, we show that the system creates highly precise markings at a relative precision of 1.5 mm and a global accuracy of 5-6 mm and discuss the results in the context of future construction robotics.}


\maketitle

\section{Introduction}
\label{sec:intro}
With the increased capabilities and success stories of autonomous systems, more and more applications previously considered impossible for robotic systems have come within reach. As a result, companies invest more heavily in developing robotic systems. Construction remains one of the least automatized industries, however, also in construction, there is a notable trend towards robotics technologies \citep{2018bcg}. Early mover companies have started developing ground robots for construction site tasks. For example, Hilti recently presented a semi-autonomous drilling robot\footnote{\href{"https://www.hilti.com/content/hilti/W1/US/en/business/business/trends/jaibot.html"}{"Semi-autonomous construction robot Jaibot"}, \textit{last accessed on 17.02.2023}}, and Husqvarna offers remote demolition robots\footnote{\href{"Husqvarna DXR remote demolition robots"}{https://www.husqvarnacp.com/uk/machines/demolition-robots/}, \textit{last accessed on 17.02.2023}}. 

The previously mentioned systems have a limited workspace. Due to their ground robot nature, they have limited work heights and need traversable, load-bearing paths to move.  Aerial robots, on the other hand, can operate at high altitudes and in areas difficult to traverse. However, they are susceptible to disturbances, increasing the challenge of performing exact and dexterous tasks. Commercially available \acp{MAV} exist for asset management and visual inspection with collision-resilient platforms\footnote{\href{https://flybotix.com/asio-drone/}{"ASIO the revolutionary indoor inspection drone solution"}, \textit{last accessed on 17.02.2023}}.
\begin{figure}[bt]
    \centering
    \includegraphics[width=\linewidth]{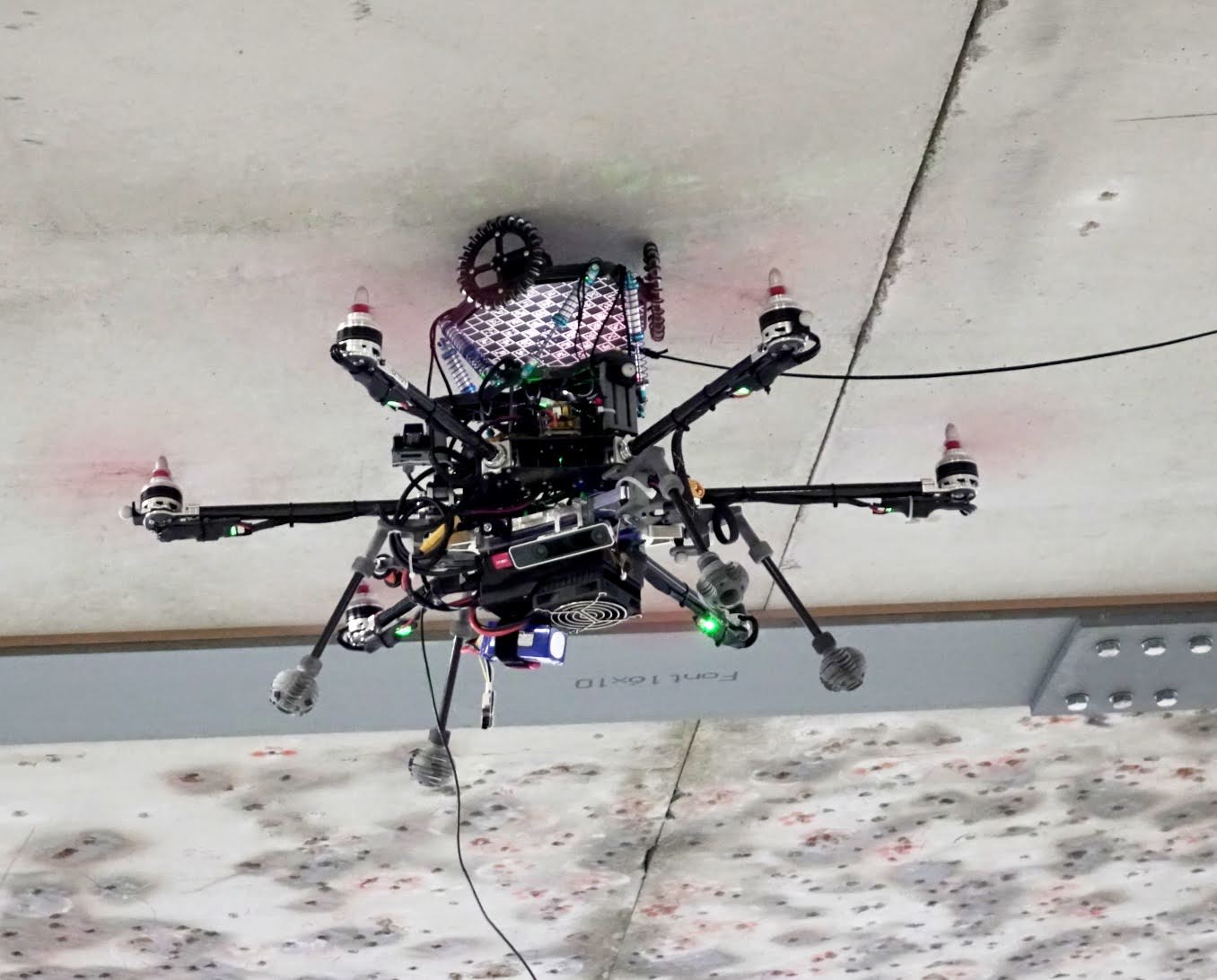}
    \caption{Our aerial robot during a layouting mission on a construction site accurately marking points on a concrete ceiling.}
    \label{fig:full_system}
\end{figure}
Tasks in construction environments require contact with the environment, a regime for which these commercial systems were not designed. A recently emerging class of aerial robots are omnidirectional \acp{MAV}, which are able to exert forces and torques in arbitrary directions in a controlled manner. Early examples are the \citep{kbodie2019omav} and the commercialized remote-controlled Voliro\footnote{\href{https://www.voliro.ch}{"Voliro AG"}, \textit{last accessed on 17.02.2023}}. However, there exist still a wealth of technical hurdles in state estimation and planning for reliable, precise, and autonomous missions on-site.
A task of interest in construction is layouting, i.e., marking points, lines, and curves on the ceiling to indicate areas where to drill or anchor components. The manual marking process is repetitive and troublesome, especially at height, and small errors accumulate, which can lead to costly remediations. The level of accuracy required from the layouting tool has yet to be achieved with common aerial robots on-site, which typically operate at centimeter accuracies.
\subsection{Related Works}
\label{ssec:relworks}
In recent years, the research community investigated the usage of aerial manipulators for non-destructive contact-based inspection and manipulation \citep{2021aerialmanip}.
Previous works have shown that aerial manipulators can accurately measure material thickness using ultrasonic probes \citep{voliroTNDE2021} or locate material defects with eddy-current sensors \citep{2019-Tognon_NDT}. Impressive progress has also been made in push-and-slide inspection, with aerial systems being able to follow curved surfaces without losing contact \citep{Nava2020DirectManipulators}. Several works improve precision and disturbance rejection using parallel manipulators. End-effector accuracy can be improved by at least an order of magnitude with the help of a delta-manipulator \citep{chermprayong2019integrated}. \cite{Tzoumanikas-RSS-20} showed impressive drawing precision in the millimeter range, jointly controlling the \ac{MAV} and delta-arm with nonlinear model predictive control.  \cite{kovac2023sensorplacement} showed in a remarkable number of experiments that also classical coplanar \acp{MAV} can be used to autonomously place sensors on planar surfaces within a few centimeters.
All the above-presented aerial robots were operating in lab conditions. However, transitioning into real applications involves robustness to environmental uncertainties and reliable decisional autonomy~\citep{2021aerialmanip}, which our work addresses. 
 
 Despite the great efforts from the research community, complete systems that can operate under more realistic conditions remain rare. \citet{2020fullyactuatedcontactinspection} demonstrated an autonomous aerial robot for bridge inspection purposes equipped with multiple sensors fused in a Kalman filter to estimate the robot's position. The authors address the problem of robustly aligning different reference frames by running a separate calibration procedure beforehand. While their sensor fusion framework can handle measurement dropouts, data collection for the calibration procedure needs to be repeated after each system restart.  Additionally, there is no information on the system’s absolute accuracy; thus, it is unclear how well the system performs after longer operation times when it accumulates drift and the statically obtained alignment no longer holds. \cite{Trujillo2019NovelIndustry} presented a semi-autonomous aerial robot for \ac{NDT} in the oil and gas industry. The robot has two operating modes, a free flight mode and a contact-mode that keeps the robot steady with respect to the contacted surface. Switching between modes is handled by the human operator. When in contact the robot's pose is estimated with respect to the contacted surface using image-based feature tracking. While this enables the aerial robot to robustly follow surfaces, the achieved accuracy (the authors claim an accuracy of \SI{1.8}{\centi\metre} per traveled meter) is not sufficient for construction-related tasks.

\subsection{Contributions}
In our previous work \citep{rsspaper}, we presented the design and evaluation of the here-used end-effector concept for aerial layouting under laboratory conditions. In this paper, we extend our previous work and take a holistic look at all challenges present in aerial layouting under real conditions. Namely, our contributions include
\begin{itemize}
    \item a complete, novel system capable of aerial layouting at height without the need for a motion capture system,
    \item a state estimation and sensing strategy for robust flight close to and in contact with structure,
    \item a novel navigation approach tailored to the unique setting of a flying robot with a spring-decoupled and independently actuated end-effector,
    \item and finally, a comprehensive experimental study of achieved accuracy and precision, and contextualization thereof in a construction setting.
\end{itemize}
To the best of our knowledge, this is the first aerial layouting system capable of precise markings on-site without the need for a motion capture system.
As this paper is an extension of a conference paper, the chapters concerning the end-effector (\cref{sec:ee_design}, \cref{sec:ee_eval}) are adapted from our previous work.

\section{Method}

\label{sec:method}
In this section, we present the main components required for high-accuracy layouting on ceilings with an aerial robot. The robot consists of two components, an omnidirectional flying base that can exert forces in any direction and a compliant and actively driven end-effector (see \Cref{fig:full_system}). 
In our previous work \citep{rsspaper}, we showed that this end-effector has been crucial for achieving millimeter precision using a flying base. In order to fully utilize the capabilities of the end-effector in real-world construction environments, robust and accurate state estimation and navigation components are needed.
We propose using a pose-graph-based state estimator with a dual-graph design to decouple global and local state estimation in combination with a highly modular Riemannian Motion Policy-based navigation system that exploits the compliant nature of the end-effector reactively.
All main components are presented in detail in the remainder of this chapter.
\subsection{Flying base}
\begin{table*}
\centering
\begin{tblr}{
colspec = {llrX[l]},
cell{2}{1-4} = {m},
}
\hline
Type & Model & Rate & Characteristics \\
\hline
IMU & Adis 16448B & \SI{200}{\hertz} & {Turn-on bias: \SI{20}{\milli\gram}, \SI[per-mode=symbol]{1800}{\degree\per\hour} \\ Random walk: \SI{187}{\micro\gram\per\sqrt{\hertz}}, \SI{0.66}{\degree\per\sqrt{\hour}}}  \\
Camera & Realsense T265 & \SI{30}{\hertz} & \\
Reflector & GRZ101 \SI{360}{\degree} Mini Prism & \SI{15}{\hertz} & \SI{1.5}{\milli\metre} pointing accuracy
\\ 
Camera & See3CAM 20CUG & \SI{60}{\hertz} & 2 MP global shutter , \textit{PT-01224XFL} lens\\
Depth Camera & PMD flexx 2 & \SI{30}{\hertz} & $224 \times 172$ px resolution, $\SI{0.1}{\metre}-\SI{4}{\metre}$ range\\
\hline
\end{tblr}
\caption{\label{tab:sensor_specs} Sensor specifications.}
\end{table*}
Our flying base is a custom-built hexacopter with servo-motor actuated propeller arms. It is possible to change the tilt angle of each arm individually and to effectively decouple the translational movement of the flying base from its orientation. The flying base uses six \textit{KDE 3510XF-475} motors with 12x4.5 propellers and RPM-controlled ESCs for propulsion. Six \textit{Dynamixel XM430-W350} control the orientation of the arms. Additionally, the robot carries an \textit{Nvidia AGX Orin} onboard computer running Ubuntu 20.04 and an \textit{ADIS16448B} \ac{IMU}. Two onboard cameras track the flying base's pose and the end-effector's relative displacement, respectively. A reflective prism locates the robot's absolute position with respect to accurate measurements from a total station. A depth camera is used to measure the relative distance to the surface of interest. \Cref{tab:sensor_specs} lists more detailed information on all sensors mounted on the flying base. The impedance controller developed by \cite{kbodie2019omav} is running on the onboard computer to control the pose of the flying base. 

\subsection{End-Effector Design}
\label{sec:ee_design}
The main goal of the end-effector is to correct the positional imprecision of the flying base mostly stemming from unforeseeable in-flight disturbances. With this in mind, our end-effector incorporates three main design choices:
\begin{enumerate}
    \item \textit{Compliance} between the end-effector tool and the flying base to decouple disturbances acting on the base and thus increasing the precision of the marking;
    \item \textit{Multiple contact points} between the ceiling and the end-effector to increase system stability and constrain the end-effector's attitude;
    \item \textit{Actuation} on the end-effector to allow controlled movement along the ceiling and further refinement of the tool position;
\end{enumerate}
With these design objectives in mind, we developed an end-effector based on a Gough–Stewart platform \citep{stewart1965platform}, a parallel manipulator with six degrees of freedom. Flight simulators commonly use this mechanism, where linear actuators control the cabin. \cite{hu20186} recently developed a passive Gough-Stewart structure by replacing the actuated legs with passive springs for vibration isolation. Similarly, our end-effector design has spring-dampers to connect the flying base with the movable end-effector tool. The design allows the flying base to push upwards and compress the spring-dampers, suppressing vibrations and effectively increasing the stability of the flying base. 

The spring-dampers are off-the-shelf \textit{Z-D0033 Dual Spring Shocks} for radio-controlled model cars.
We adjusted the spring stiffness such that a force of \SI{15}{\newton} halves the distance between the end-effector and the flying base. The required force is a trade-off between available thrust, ceiling grip and compliance. The damping fluid is water. By experimentation, it was the only fluid that allowed the upper platform to return to its maximal height with the given spring stiffness. The end-effector's geometry is optimized such that it can be displaced by up to \SI{2}{\centi\metre} in directions parallel to the flying base. The displacement is sufficient to compensate for in-flight disturbances acting on the flying base. The typical flight control mean absolute error in all directions is less than \SI{2}{\centi\metre}, as displayed in \Cref{tab:design-validation-results}.

The end-effector is equipped with three custom-built omni-wheels to provide multiple contact points while still permitting smooth movement in any direction. Each wheel is actively driven by a servo-motor in order to compensate for tracking errors. The servos have a velocity controller that tracks a reference velocity provided by the navigation algorithm. The end-effector is additionally holding a retractable permanent marker for marking purposes. An upward-facing camera tracks the relative displacement between the flying base and the end-effector. It faces a \textit{ChArUco} board attached to the end-effector's bottom side. The computer vision library \cite{2015opencv} provides the \textit{ChArUco} tracker. The accuracy of the tracking system was evaluated in a static experiment using a Vicon motion capture system, reaching a tracking error of \SI{0.8}{\milli\metre} in position and \SI{0.2}{\degree} in yaw on average\citep{rsspaper}.

\subsection{Frame Definitions}
\begin{figure}[bt]
    \centering
    \includegraphics[width=\linewidth]{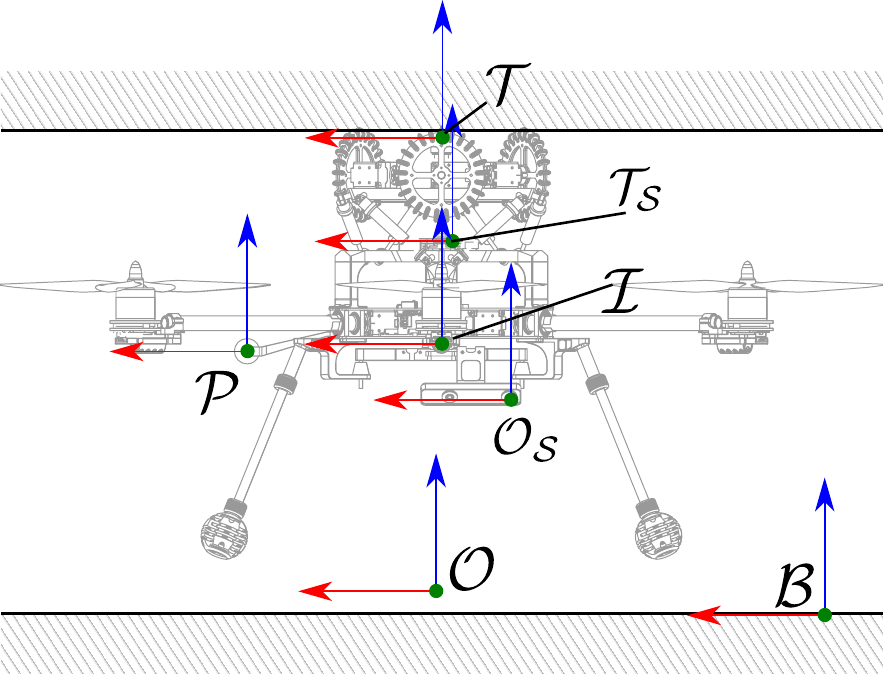}
    \caption{Overview of the used coordinate frames. Nomenclature according to \cref{tab:frame_definitions}.}
    \label{fig:coords}
\end{figure}

The entire system has seven different coordinate frames, shown in \Cref{fig:coords}.
In the following, we describe the frames, their relation to each other, and the determination of their transformations. 

 The target locations are given with reference to the globally fixed building frame~$\fB$. The body-fixed frame~$\fI$ is located at the \ac{IMU}'s point of percussion. It defines the pose of the aerial robot with reference to the odometry frame~$\fO$, which in our case, is the initial starting position of the robot. The location of the \ac{VIO} sensor and the prism is denoted as $\fOs$ and $\fP$, respectively, and the fixed transformation from the \ac{IMU} to these sensors was obtained from the robot's CAD model. The tool sensor frame $\fTs$ is located at the end-effector camera, and its relative pose with respect to the IMU was obtained from an extrinsic calibration beforehand. The pose of the end-effector tool $\fT$ relative to $\fTs$ is estimated through the end-effector camera tracking the \textit{ChArUco} board. 

The homogenous transformation from the odometry to the \ac{IMU} frame $\Toi \in SE(3)$ is estimated using only relative measurements provided by the onboard sensors. As a result, the estimated transform drifts with time. To account for this drift we additionally estimate the transformation from the building to the odometry frame $\Tbo$. 
\Cref{tab:frame_definitions} summarizes the abbreviation of all frames, their parent frames, and how the transformation to their parent frame was obtained.
\begin{table}[bt]
\centering
\begin{tblr}{
colspec = {r|c|c|c|c},
cell{3,5}{1-5} = {r=2}{m}
}

Name & Abbr. & Parent & Fixed & Source \\
\hline
Building   & $\fB$  & -   & \cmark & - \\
Odometry   & $\fO$  & $\fB$ &        & {Global \\ estimate} \\
& & & & \\
IMU        & $\fI$  & $\fO$ &        & {Local \\ estimate} \\
& & & & \\
VIO sensor & $\fOs$ & $\fI$ & \cmark & CAD model \\
Prism      & $\fP$  & $\fI$ & \cmark & CAD model \\
Tool sensor& $\fTs$ & $\fI$ & \cmark & Calibration \\
Tool       & $\fT$  & $\fTs$&        & Tracking \\
\end{tblr}
\caption{\label{tab:frame_definitions} Description of all coordinate frames, their relation, and determination source.}
\end{table}

\subsection{Robust Sensor Fusion}
\label{sec:robustsensorfusion}
\begin{figure*}
\includegraphics[width=\linewidth]{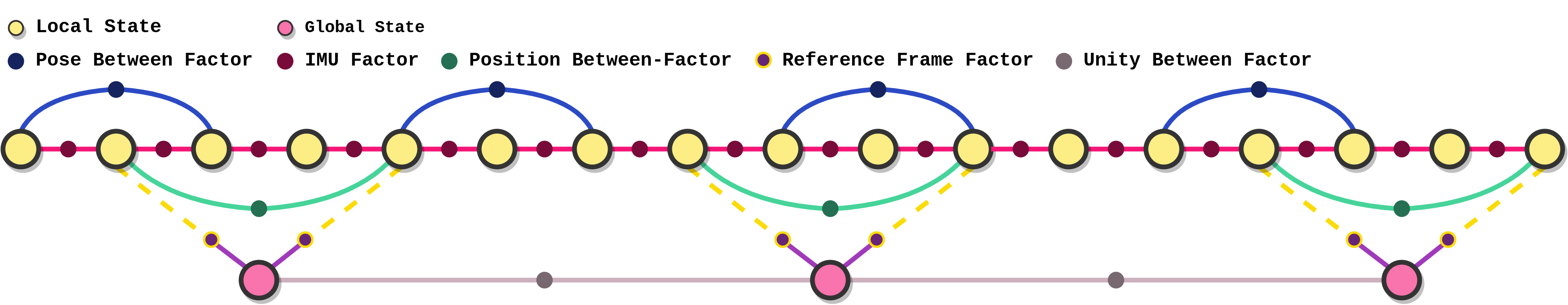}
\caption{Visualization of the dual-graph structure utilized in our state estimation framework. The local states are constrained purely by relative measurements in order to robustly handle measurement dropouts. The global graph independently infers the global pose using the local state estimate and position measurements provided by the total station.}
\label{fig:factor_graphs}
\end{figure*}
Construction surveyors typically use total stations to stake out points of interest referenced to a globally fixed building frame $\fB$. Robotic total stations can additionally track moving reflectors. Thus we use a total station to locate our robot. Unfortunately, the position of the prism is insufficient. We require a high rate estimate of $\Tbt$, which describes the pose of our marking tool within $\fB$. The continuous estimate can be obtained by fusing multiple sensors and co-estimating their drifting calibrations.

Most approaches try to estimate the full global state of the system in a single filtering or smoothing framework \citep{msf2013, confusion2019, INDELMAN2013721}. However, systems with dense kinematic chains, multiple measurements, and states are highly non-linear and inherently hard to tune. Too many correlated states, some of which are not directly observable, need to be co-estimated and cause the optimization to get stuck in local minima. Furthermore, filtering methods are susceptible to outliers and measurement dropouts, causing them to become overconfident or diverge.

When tracking dynamic objects like aerial vehicles, total stations often lose track of the prism. With this in mind, we developed a pose-graph-based sensor fusion algorithm similar to \cite{nubert2022graph}, based on a dual-graph design. In their work, the authors addressed the problem of measurement dropouts by switching between two different optimization problems depending on the availability of global pose measurements. We, on the other hand, propose two loosely coupled optimization problems, depicted in \Cref{fig:factor_graphs}. 

The first optimization performs inference over a factor graph consisting only of states constrained by relative measurements. The relative nature of the constraints allows temporary dropouts and subsequent returns of measurements but comes at the cost of accumulating errors over time, causing the state estimate drift. Controlling the flying base requires only a locally consistent and smooth state estimate. The local factor graph is structured to favor robustness and local consistency over global estimate accuracy.

A second optimization uses the local state estimates and position measurements from the total station to infer the drifting pose of the odometry ($\fO$) frame origin with respect to $\fB$. The global pose estimate is only used for end-effector navigation. Therefore, if measurements from the total station are lost during the operation, only the system's global accuracy is affected. The aerial robot itself, however, remains fully operational.

The sensor fusion algorithm is implemented using the GTSAM framework \citep{gtsam}. In the following, we present the structure of the two different graphs in more detail.
\subsubsection*{Local State Estimation}
The local state of the aerial robot at a time t is defined as
\begin{equation}\label{eq:local_state}
{}_{\fI}\boldsymbol{x} \coloneqq
\left[
\symbolB{q}{\fO\fI} , \symbolA{\fO}{p}{\fI} , \symbolA{\fO}{v}{\fI}, \symbolA{\fI}{b}{a}, \symbolA{\fI}{b}{g}
\right]
\end{equation}
with $\symbolB{q}{\fO\fI} \in SO(3)$ being the rotation from $\fO$ to $\fI$. $\symbolA{\fO}{p}{\fI} \in \mathbb{R}^3$ and $\symbolA{\fO}{v}{\fI} \in \mathbb{R}^3$ represent the position and velocity of $\fI$ relative $\fO$. $\symbolA{\fI}{\mathbf{b}}{a} \in \mathbb{R}^3$ and $\symbolA{\fI}{\mathbf{b}}{g} \in \mathbb{R}^3$ are the biases of the \ac{IMU}'s accelerometer and gyroscope modeled as integrated white noise expressed in $\fI$. 

Every \ac{IMU} measurement adds a new state ${}_{\fI}\boldsymbol{x}$ to the local graph. Each of these states is connected to their previous one by an \textit{\ac{IMU} factor}. Every \ac{IMU} factor is included in the optimization cost as the following additive term

\begin{equation}
 \left\| 
 \symbolB{r}{\mathrm{I}} 
 \right\|_{\Sigma_{\mathrm{I}}}^2 \quad , \mathrm{with} \quad \symbolB{r}{\mathrm{I}} \coloneqq \left[ \symbolB{r}{\Delta\mathbf{R}}^\mathrm{T} , \symbolB{r}{\Delta\boldsymbol{v}}^\mathrm{T} , \symbolB{r}{\Delta\boldsymbol{p}}^\mathrm{T} \right]^\mathrm{T}
\end{equation}
with covariance $\Sigma_{\mathrm{I}}$. $\symbolB{r}{\Delta\mathbf{R}}$, $\symbolB{r}{\Delta\boldsymbol{v}}$, and $\symbolB{r}{\Delta\boldsymbol{p}}$ are residual errors of relative motion increments in orientation, velocity, and position, as described by \citet[Equation 45]{forster2016manifold}. Odometry measurements from the \ac{VIO} sensor and position measurements from the total station are added as \textit{between factors} to the local graph. These represent the error between the predicted and measured relative displacement in pose or position. Both are added at the sensor rate stated in \Cref{tab:sensor_specs} and always connected to the two states closest in time to the measurements.

The state estimate from the local pose graph optimization is directly used in the controller of the flying base. Therefore a high-frequent state estimate is required. We optimize the pose graph at \SI{30}{\hertz} and use a fixed-lag smoother with a relatively small window size of \SI{0.5}{\second}, trading better accuracy for lower computational cost. In between optimizations, the state is integrated using \ac{IMU} measurements and provided to the controller at \SI{200}{\hertz}.

\subsubsection*{Global State Estimation}
The global state
\begin{equation}\label{eq:global_state}
{}_{\fO}\boldsymbol{x} \coloneqq
\left[
\symbolB{q}{\fB\fO} , \symbolA{\fB}{p}{\fO}
\right]
\end{equation}
defines the drift in orientation $\symbolB{q}{\fB\fO}$ and position $\symbolA{\fB}{p}{\fO}$ of $\fO$ wrt. $\fB$. The state estimate of the local graph and position measurements from the total station are matched based on their timestamp and added as a \textit{reference frame factor} to the graph. This factor is included in the optimization as the following position residual 
\begin{equation}
\left\| \boldsymbol{r}_\mathrm{G} \right\|_{\Sigma_{\mathrm{G}}}^2
=
     \left\| 
 \symbolA{\fB}{p}{\fP} - 
 \left( \symbolA{\fB}{p}{\fO} + \symbolB{q}{\fB\fO}\left(\symbolA{\fO}{p}{\fP} \right)\right)
 \right\|_{\Sigma_{\mathrm{G}}}^2
\end{equation}
with covariance $\Sigma_{\mathrm{G}}$, the measurement from the total station $\symbolA{\fB}{p}{\fP}$ and $\symbolA{\fO}{p}{\fP}$  obtained from the pose estimate of the local graph
\begin{equation}
    \symbolA{\fO}{p}{\fP} =  \symbolA{\fO}{p}{\fI} + \symbolB{q}{\fO\fI}\left(\symbolA{\fI}{p}{\fP} \right).
\end{equation}
The position of the prism with respect to the \ac{IMU} $\symbolA{\fI}{p}{\fP}$ is obtained from the CAD model of the system. 

As this factor provides a residual solely on the position, the orientation $\symbolB{q}{\fB\fO}$ is only indirectly observable through the lever arm between $\fI$ and $\fP$ (see \Cref{fig:coords}) and sufficiently distinctive measurements. Hence, the global state ${}_{\fO}\boldsymbol{x}$ is only observable when the aerial robot is in motion. We, therefore, only add new states with every 15th measurement to the graph, corresponding roughly to one state per second.
A unity \textit{between factor} enforces consistency between two consecutive global transformation states. The identity assumption acts as a prior for the expected drift.

The rate at which the local state estimate accumulates drift is slow compared to the rest of the system's dynamics. We, therefore, optimize the global pose graph only at \SI{5}{\hertz}. This allows us to optimize over a larger sliding window of \SI{20}{\second} while keeping the computational cost low.

\subsection{Reactive Riemannian navigation}
The flying base and the end-effector are controlled independently and linked through a compliant suspension. While necessary for accuracy, this design adds several unusual degrees of freedom, e.g., the body-tool offset, which we define as the distance and relative orientation between the flying base and the end-effector. Due to the physical nature of aerial interaction, unpredictable forces affect the system, such as airflow disturbances, irregular contact, or ceiling effects. 
To address these unique challenges, we propose to use a modular and reactive navigation architecture based on Riemannian Motion Policies \citep{ratliff2018riemannian}. The well-defined geometric-mathematical structure of \acp{RMP} allows the formulation of motion constraints in multiple different, potentially diverging, coordinate frames. Compared to optimization- or sampling-based algorithms, we do not need to commit to a single global and consistent overall state estimate and can stay robust in the presence of drift and sensing inaccuracies. The reactive nature of \acp{RMP} is ideal for coping with the unpredictability of aerial manipulation and layouting. 

\subsubsection*{Riemannian Motion Policies}
In the following, we summarize the most important characteristics and operators of \acp{RMP}. Please refer to \citep{ratliff2018riemannian} for more details.
The main idea behind \acp{RMP} is to decouple a navigation problem into small individual policies defined in the task manifold $\Mtask$ with dimensionality $\dtask$ where a given problem is easiest to solve. Each policy defines a state-dependent acceleration $f \in \mathbb{R}^{\dtask}$ and Riemannian metric $A \in \mathbb{R}^(\dtask\times\dtask)$, which are then locally mapped to a common configuration manifold $\Mconf$ (e.g. $SE(3)$). The metric allows the weighting of individual policies, relative to others, directionally or axis-wise. 
The mapping, called the pull-back operator, from manifold $\Mtask$ to $\Mconf$ of the policy $(f,A)$ is defined as
\begin{equation}
    \textit{pull}((f,A)_{\Mtask}) = ((J^{T}AJ)^{+}J^{T}Af, J^{T}AJ)_{\Mconf},
    \label{eq:rmp_map}
\end{equation}
where $J$ is the local Jacobian relating the two manifolds at a given position.
The mapped policies add up into a single metric-weighted sum according to \Cref{eq:rmp_sum}. 
\begin{equation}
    \textit(\Bar{f},\Bar{A}) = \left(\left(\sum_{i}A_{i}\right)^+ \sum_{i}A_{i}f_{i}, \sum_{i}A_{i}\right)
    \label{eq:rmp_sum}
\end{equation}
Finally, the robot executes the resulting acceleration $\Bar{f}$.

We structure the overall navigation approach into five policies that drive the complete system in a coordinated, robust, and exact manner. \Cref{tab:policies} provides an overview of the different policies, their metrics, and operation manifolds.
The general mode of operation is to approach the ceiling through a depth-servoing policy and push against the wall to provide friction for the end-effector by activating a spring-loading policy. Once firmly in contact with the ceiling, the end-effector navigation policy aggressively drives the end-effector to the desired target. Due to the mechanical design, the range of independent movement of the end-effector is constrained to about $\SI{2}{\centi\metre}$. Thus, the end-effector following policy continuously centers the flying base below the end-effector.  To summarize, in free flight, the flying base operates as a normal aerial robot, but once in contact, the end-effector operates independently, and the flying base follows it to provide stability. Finally, the prism tracking policy constrains the platform yaw such that the prism remains in direct line-of-sight to the total station. Generally, we consider our configuration manifold $\Mconf$ to be $SE(3)$ for the flying base and $SE(2)$ for the end-effector.
\begin{table*}[tb]
\begin{tabular}{|p{0.165\linewidth}|p{0.15\linewidth}|p{0.12\linewidth}|p{0.12\linewidth}|p{0.15\linewidth}|p{0.12\linewidth}|}
\hline
Policy     & End-effector navigation & Spring loading & Depth servoing & End-effector following & Prism tracking    \\ \hline
Task Manifold & $\fT$                  & $\fTs$         & $\fI$           & $\fTs$                 & $\fI$              \\
$diag(A)$      & $[1,1,0,0,0,0]$            & $[0,0,1,0,0,0]$    & $[0,0,1,0,0,0]$    & $[1,1,0,0,0,0]$           & $[0,0,0,0,0,1]$ \\
Conf. Manifold & $\fB$                  & $\fI\rightarrow\fO$         & $\fO$           & $\fI\rightarrow\fO$                 & $\fO$              \\
Executed on   & EE                      & FB             & FB             & FB                     & FB                \\ \hline
\end{tabular}
    \caption{Overview of the used policies. Task manifold is the space in which the policies $f$ function is defined, whereas configuration manifold is the space the policy is mapped into for execution. Some policies are mapped through a chain of spaces, indicated by an arrow. The last row indicates on which part of the robot the policy is executed. EE stands for end-effector and FB for flying base. For the formulation of the $f$ function of each policy, please refer to the supplementary material (Online Resource 2). The axis order in the metric matrix is $x,y,z,$roll,pitch,yaw.}
    \label{tab:policies}
\end{table*}

Most policies follow the generic attractor scheme from \citep{ratliff2018riemannian}, which defines $f$ as an acceleration based on a soft-maxed error function.
\begin{equation}
	f = \alpha \cdot \mathbb{S}(x - x_{0}) - \beta
\end{equation}
\noindent where $\alpha$ and $\beta$ are tuning parameters, and $\mathbb{S}$ is the soft-normalization function
\begin{equation}
    \mathbb{S}(z) =\frac{z}{|z| + \gamma\ log(1+exp(\gamma |z|))}
\end{equation} with tuning parameter $\gamma$. The corresponding metric $A$ is often constructed as a non-directional diagonal matrix where individual axes can be weighted.
The following subsections give an intuitive explanation of each policy's function $f$. We provide a comprehensive set of equations to facilitate reproducibility in the supplementary material (Online Resource 2).
\subsubsection*{End-effector navigation}
The end-effector navigation policy is an attractor that, when activated, drives the end-effector to the specified marking target location in the building frame $\fB$.
The current position of the end-effector is calculated through a concatenation of all necessary transforms from $\fB$ to $\fT$. This policy is executed independently of the flying base on the end-effector wheels.
\subsubsection*{End-effector following}
A simple $2d$ attractor that drives the flying base to be exactly below the independently moving end-effector.
\subsubsection*{Depth servoing}
A $1d$ attractor moves the flying base towards the ceiling until the end-effector is in contact. This policy operates based on the output of a time-of-flight depth camera.
\subsubsection*{Spring loading}
The spring policy exploits the underlying impedance controller to drive the flying base upwards until a desired spring extension is reached. The end-effector spring extension can be measured based on the estimate between $\fT$ and $\fTs$. This is equivalent to controlling the pushing force without needing a force sensor. For safety, the metric of the spring loading policy decays exponentially if the desired spring load is exceeded.
\subsubsection*{Prism tracking}
The prism tracking policy influences the yaw axis based on the current distance between the prism and an imaginary, gravity-aligned plane spanned by the total station and the flying base center. Keeping the prism in this plane ensures visibility.
\\

Each policy can be disabled by temporarily multiplying the corresponding metric with zero. We use simple state-based rules or operator buttons to enable individual policies. All policies also have tuning parameters. However, tuning has proven to be robust and can be done for each policy independently, simplifying the tuning process significantly.
As all involved manifolds are equivalent to $SE(3)$, the Jacobians between them are simple rotation matrices.

The evaluations and summations of policies are executed at the controller frequency of \SI{200}{\hertz}, with a CPU usage below $10\%$ of one core. Compared to a sampling-based or optimization-based planner, our proposed stack is able to react fast and without delay to any disturbance or deviation at negligible compute cost. Due to the decomposition in multiple policies, each behavior can be tested and executed independently. Except for the end-effector navigation policy, which can be disabled if needed, all policies seamlessly cope with drifting odometry due to their body-frame formulations.
\section{Results}
\label{sec:results}
We evaluate individual parts as well as the complete system, with a focus on accuracy, precision, and robustness. The contribution of the mechanical end-effector design is evaluated through a set of flight experiments under simplified conditions. The most important characteristics of the state estimation and navigation algorithms are evaluated individually. Finally, the complete system is demonstrated, and its precision and accuracy are evaluated in a laboratory setting and on a construction site.
For the final evaluation, only x and y position errors, i.e., errors in the plane parallel to the ceiling, were considered as the ceiling constrains the end-effector in height and attitude.
\subsection{End-effector}
\label{sec:ee_eval}
To evaluate the end-effector design, we performed an ablation study in which we removed individual features from the end-effector and compared the precision of the different configurations. During the experiments, the aerial robot followed a circular trajectory with a radius of \SI{250}{\milli\metre}. The maximum velocity and acceleration were limited to  \SI[per-mode = symbol]{5}{\centi\metre\per\second} and \SI[per-mode = symbol]{2.5}{\centi\metre\per\second\squared}, respectively. All experiments were conducted using a Vicon motion capture system, providing accurate and non-drifting pose estimates used by the controller of the aerial robot. This removes error sources unrelated to the end-effector design. The results presented in \Cref{tab:design-validation-results}, therefore, represent the maximum achievable precision of the system. 

\begin{table*}[bt]
\begin{tblr}{
colspec = {|rc|c|c|c|c|X[c]|X[c]|X[c]|X[c]|X[c]|X[c]|},
cell{1}{7,10} = {c=3}{c},
cell{1}{1-6}  = {r=2}{f},
cell{2}{7-12}    = {r=1}{f},
vline{4-6} = {dotted}
}

 \hline
  Experiment & \# &
  \rotatebox{90}{Contact}
 & \rotatebox{90}{Wheels} & \rotatebox{90}{Compliance} &  \rotatebox{90}{Actuation} & {\\ End-effector \\ xy-error [mm]} & & & {\\ Flying Base \\ xyz-error [mm]} & & \\ 
  & & &  &  &  &  
 MAE & STD & P\textsubscript{90\%} &
 MAE & STD & P\textsubscript{90\%} \\
 \hline
 Free-Flight &
 \ding{172}          &        &        &        &         
                     & 22.6  & 13.0   & 43.5   & 21.  6 &  9.0 & 33.8 \\

 \hline[dashed]
 Friction-less &
 \ding{173}          & \cmark & \cmark &        &          
                     & 14.5 &  7.9   & 25.0   & 13.2   &  5.8 & 20.8 \\
 Spring-Dampers &
 \ding{174}          & \cmark & \cmark & \cmark &         
                     & 33.1  & 15.3   & 54.5   & 15.4   &  6.2 & 23.6 \\
 Actuation &
 \ding{175}          & \cmark & \cmark &        & \cmark   
                     &  1.3  &  0.8   &  2.4   &  \textbf{7.7} &  \textbf{1.5} &  \textbf{9.8} \\
 Full System &
 \ding{176}          & \cmark & \cmark & \cmark & \cmark
                     &  \textbf{1.0}   &  \textbf{0.5}   &  \textbf{1.7} & 12.2 &  3.4 & 15.6 \\
 \hline
\end{tblr}
\caption{\ac{MAE}, standard deviation, and the 90th percentile for the end-effector and aerial vehicle for every individual design validation experiment.}
\label{tab:design-validation-results}
\end{table*}
As a baseline, the aerial robot tracked the reference trajectory in free-fight, reaching an average precision of \SI{22.6}{\milli\metre}. This corresponds roughly to the tracking performance of the flying base. 
To remove the compliance and actuation of the end-effector one at a time, we used rigid rods and dummy servos with frictionless bearings. By adding frictionless contact points between the ceiling and the end-effector (experiment \ding{173}) the aerial robot can almost halve the tracking error. This demonstrates that contact points with the manipulated surface can be exploited to stabilize the flying base. However, adding compliance to such a system is not beneficial, as \ding{174} shows. The stiffness of the end-effector's compliant structure is not large enough to keep the end-effector above the flying base, and without additional actuation, the end-effector just gets dragged behind.
Surprisingly, the main performance gain can be obtained by using a rigid end-effector with multiple actuated contact points, improving the precision of any previous experiment by order of magnitude (experiment \ding{175}). In addition to the multiple contact points stabilizing the flying base, the actuated wheels increase friction in unwanted directions while still keeping the friction low in the direction of the trajectory. These results also suggest that compliance is not strictly necessary to reach high precision. Nevertheless, adding compliance to the end-effector still reduces the tracking error, most likely because the end-effector can better compensate for orientation errors of the flying base (experiment \ding{176}).
All subsequent experiments do not use the motion capture system.
\subsection{Sensor fusion}
\begin{figure}[bt]
    \centering
    \includegraphics[width=\linewidth]{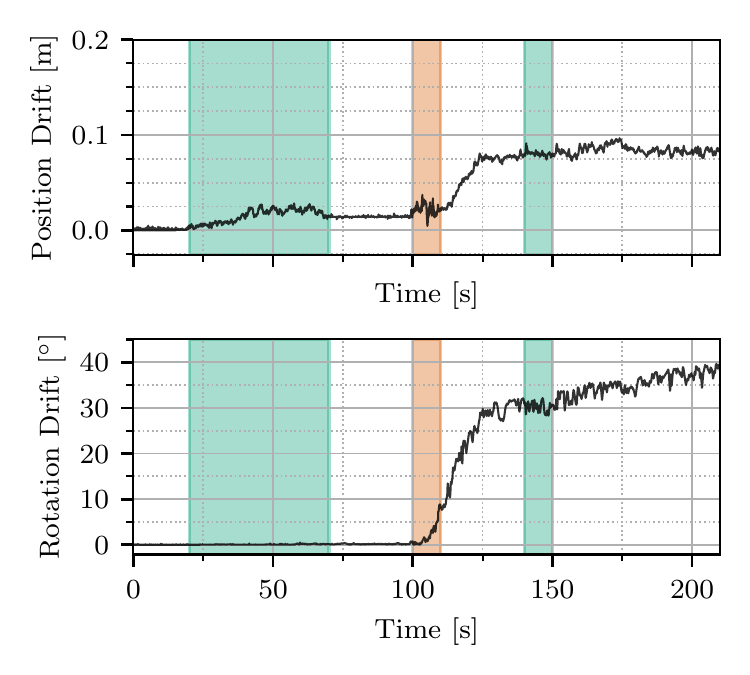}
    \caption{ The difference in absolute position and angle of rotation between a local state estimator without measurement dropouts and one with. The green and orange bars represent the loss of position and pose measurements, respectively. }
    \label{fig:local_graph_drift}
\end{figure}

This section presents a more qualitative evaluation of our sensor fusion framework. We tested the robustness of our local state estimation against sensor dropouts in an offline experiment and studied the convergence rate of our global state estimator. For a thorough evaluation of the system's accuracy, the reader is referred to \Cref{sec:full_system_results}.
The robustness of the local state estimator was evaluated by artificially removing sensor measurements that were recorded during an actual layouting experiment. In \Cref{fig:local_graph_drift} the output of the local estimator using this sparsified data is compared to an estimator provided with the full set of measurements. Green and orange bands represent periods in which position, respectively, pose measurements were missing completely.  
From the graphs, it is visible that the state estimator can handle measurement dropouts. While slowly accumulating some drift, the local estimate stays locally consistent in both position and rotation and does not jump when sensor measurements return. It shows that the local estimator can provide the required robustness for operating aerial robots under realistic conditions where sensor measurements are not constantly available. It is worth noting that the loss of pose measurements results in a more considerable drift, especially in orientation. In such cases, the orientation is estimated purely by \ac{IMU} measurements rendering the yaw unobservable and causing the orientation and position to drift slowly ($\approx$ \SI[per-mode=symbol]{1.2}{\degree\per\second} and \SI[per-mode=symbol]{0.004}{\metre\per\second}).
While errors in the global estimate do not affect the operability of the aerial robot, they do directly impact on the system's absolute accuracy.
A well-converged global estimate is, therefore, crucial for accurate end-effector positioning. Due to the nature of the available measurements, only the position and attitude of the global transform $\Tbo$ are observable. The yaw angle can only be indirectly observed from motion and, thus, is usually the largest source of error.
\begin{figure}[bt]
    \centering
    \includegraphics[width=\linewidth]{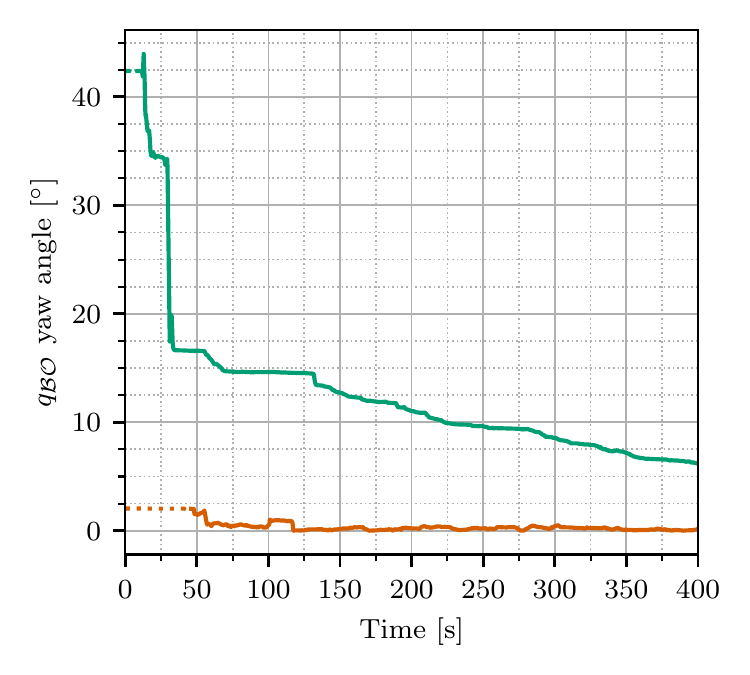}
    \caption{ The yaw angle estimate of the local odometry frame for two different experiments, one with a good initial guess (orange) and the other with a bad one (green). The dotted line indicates the time before take-off.}
    \label{fig:yaw_angle}
\end{figure}
\Cref{fig:yaw_angle} shows that the global estimator can converge to a consistent solution in yaw. However, if the initial value is wrong by a significant amount, sufficient motion and time are needed to achieve convergence. The setup time and convergence rate could be further improved by parameter tuning or by providing better initial guesses.
\subsection{Navigation}
\label{sec:results_navigation}
\begin{figure}[ht]
    \centering
    \includegraphics[width=\linewidth]{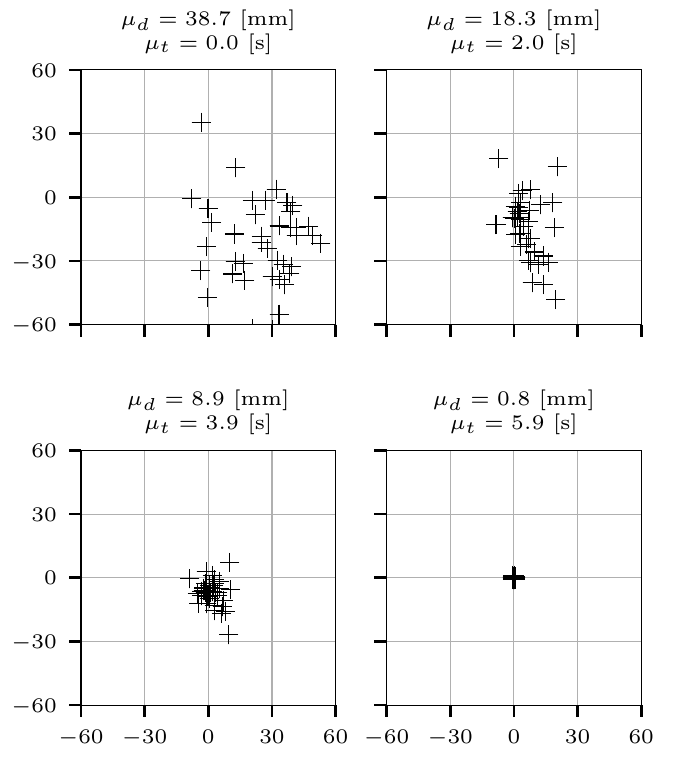}
    \caption{Deviation of the tool tip to the believed target location according to state estimation, in global frame millimeters. Top left is before enabling the policy, bottom right is at the moment of pen actuation. The remaining two panels correspond to $33.3\%$ resp. $66.6\%$ time passed between start to marking. $\mu_{d}$ is the average deviation and $\mu_{t}$ the average time since enabling the policy.}
    \label{fig:ee_policy}
\end{figure}

Amongst all policies, the end-effector navigation policy has the largest influence on the final marking accuracy.
In order to evaluate the convergence of the policy, we obtain $40$ in-flight time series, starting when the end-effector is in firm contact, and the policy is activated and ending at the release of the marker pen.
\Cref{fig:ee_policy} visualizes the millimeter $xy$ plane deviations from the tool to the target location, as estimated by the state estimator for all tries, at four different times from policy activation to marking.
The first plot corresponds to the situation when the policy is enabled -- basically, the free-flight error after attaching and stabilizing to the ceiling (\SI{38.7}{\milli\metre}). As is visualized, the end-effector policy drove the tool location to below a millimeter deviation for all tries, taking $5.9 \si{\second}$ on average. The friction between the end-effector wheels and the ceiling varied between trials. On some surfaces, such as smooth paper, there was a significant amount of wheel slippage, which posed no problem to the reactive planning methodology. 

The deviation reported in this experiment is seen from the robot in flight, i.e., the deviation to the target as output by state estimation. As the tool always converged to less than $\SI{1}{\milli\metre}$ error as estimated by state estimation, most of the global accuracy error we present in the next section comes from state estimation itself and calibration.

\subsection{Full system}
\label{sec:full_system_results}
We perform a complete end-to-end accuracy study of the whole system. No external evaluation system, such as the VICON mocap system, could  satisfactorily provide sub-millimeter accurate repeatable measurements in realistic conditions. Instead, we test the accuracy of single-point markings and the relative precision of a square grid of markings using pen on paper.
An A3 paper equipped with pre-printed fiducials is mounted to a rigid, flat ceiling, and its absolute center spot location is determined using a total station (Nova MS60, Leica Geosystems). We evaluated the repeatability of the measurement to be below \SI{1}{\milli\metre}. The aerial robot is then commanded to mark the center spot and a pattern of 4 markings in a \SI{100}{\milli\metre} $ \times $ \SI{100}{\milli\metre} square while using the total station as external tracking input as described \Cref{sec:robustsensorfusion}. The pen release mechanism is tuned to draw as small and precisely locatable points as possible. 
\begin{figure}[ht!]
    \centering
    \includegraphics[width=\linewidth]{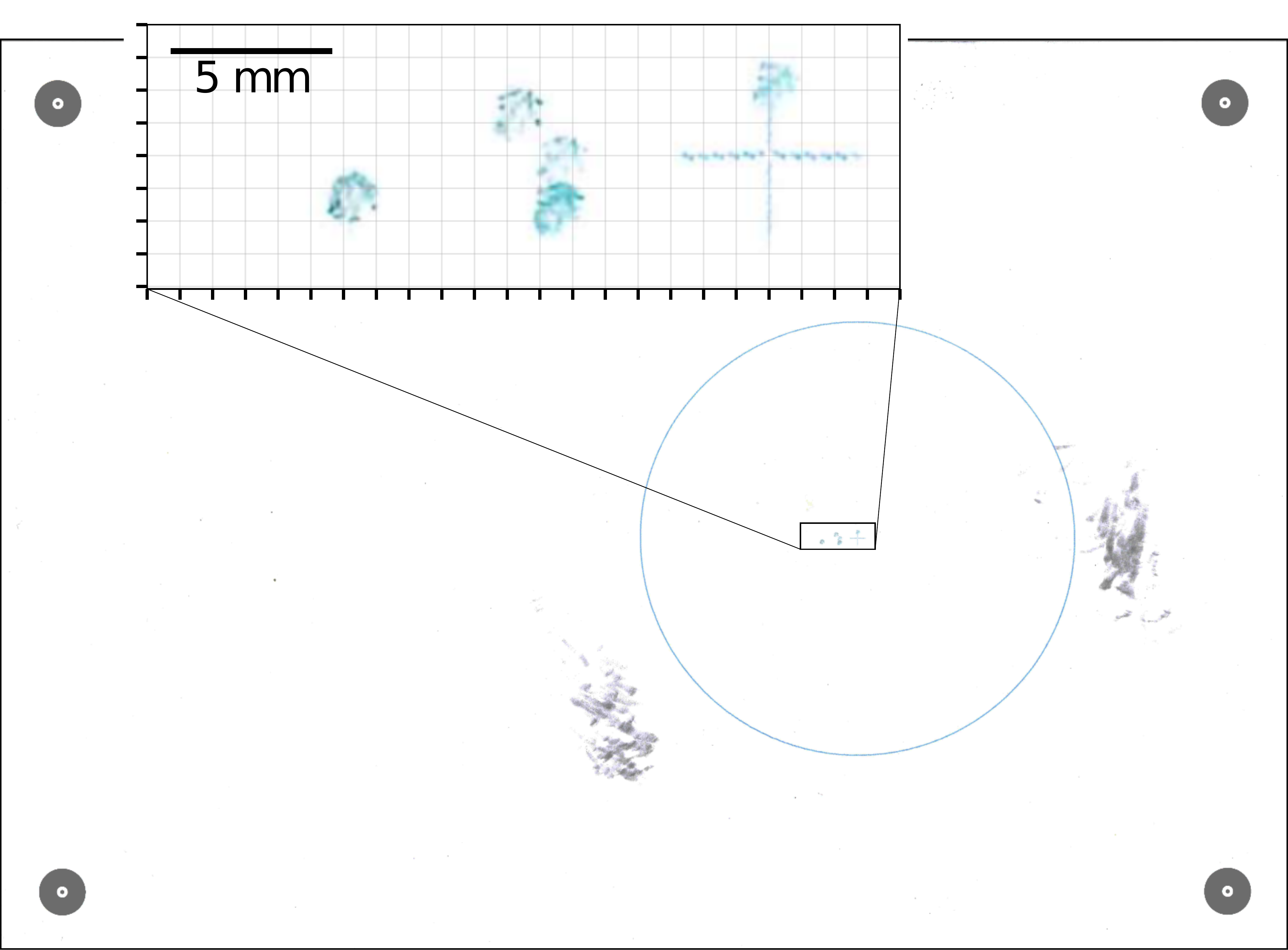}
    \caption{Example of fiducials and results on A3 paper. The magnified rectangle shows the rectified cropped image with markings and the pre-printed center point cross. Grid spacing is $1\si{\milli\metre}$.}
    \label{fig:scan}
\end{figure}
After the experiment, we scan the paper with markings at $600$ DPI, corresponding to a resolution of $\approx 26.3$ pixels per mm. Furthermore, we rectify the scanned image using a homography calculated based on the pre-printed, known fiducials and thus obtain a metrically accurate, distortion-free representation of the marked points.

\begin{figure}[bt]
    \centering
    \includegraphics[width=\linewidth]{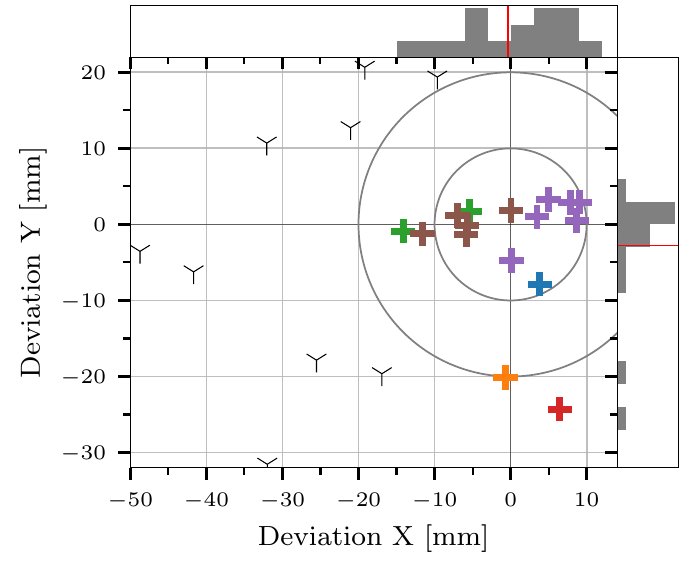}
    \caption{In-plane global accuracy deviations of markings, colored by flight. The state estimator was reset between flights, leading to slightly different converged states. To illustrate the effect of calibration, the Y-shaped markings show markings affected by slightly suboptimal camera calibrations or prism-imu calibrations.}
    \label{fig:abs_acc}
\end{figure}

\begin{figure}[bt]
    \centering
    \includegraphics[width=\linewidth]{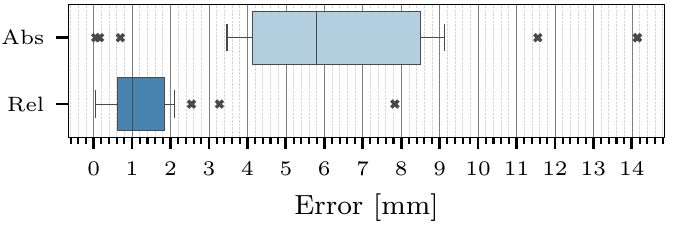}
    \caption{Comparison and statistics of end-to-end absolute accuracy (Abs) and relative precision (Rel).  }
    \label{fig:acc_comparison}
\end{figure}

\Cref{fig:scan} shows an example scan of one experiment.
We estimate the resolution of the absolute accuracy evaluation to be on the order of $\approx$\SI{0.5}{\milli\metre}, while the relative precision evaluation is accurate to within $\approx$\SI{0.04}{\milli\metre}.
\Cref{fig:abs_acc} visualizes the results of the absolute accuracy tests performed. Of the marked points, $75 \%$ are within a \SI{10}{\milli\metre} radius around the true position, with an average error of \SI{6.36}{\milli\metre}. The achievable global accuracy, as visualized in \Cref{fig:acc_comparison}, is between $5 - 6$\si{\milli\metre}. 

We attribute most of the remaining error to state estimation and calibration inaccuracies. As discussed in \Cref{sec:results_navigation}, the end-effector policy could always drive the end-effector to what the robot believes to be the true absolute global position. 
Especially the global yaw estimate has a large influence on accuracy, as it is not directly measurable -- an estimation error of \SI{1}{\degree} yields a global marking error of $\approx$\SI{3.5}{\milli\metre}. The yaw error also likely explains the larger variance of errors in the $x$ direction in \Cref{fig:abs_acc}, as it roughly corresponds to yaw misalignment in this experimental setup. 

\begin{figure}[bt]
    \centering
    \includegraphics[width=\linewidth]{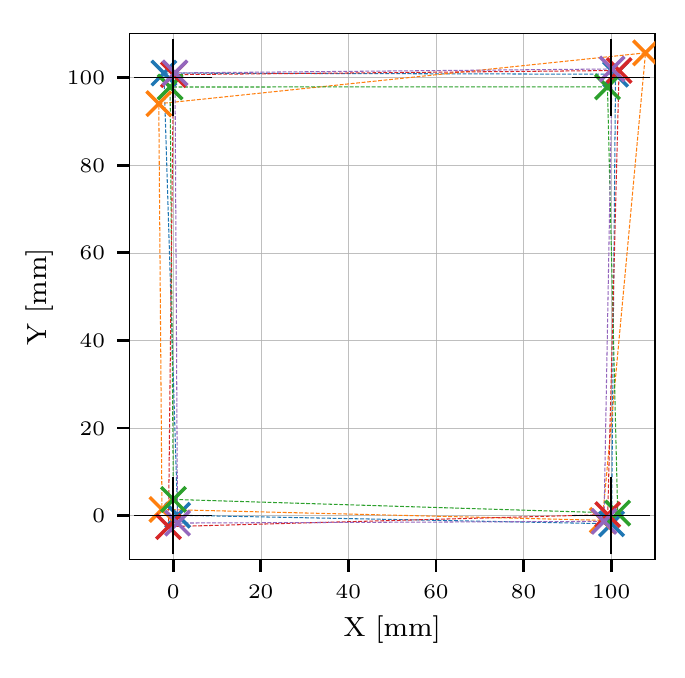}
    \caption{Visualization of five repetitions of aligned square markings, colored by repetition. One try has been omitted as one marked point was not visible on paper, likely due to mechanical jamming of the pen release. }
    \label{fig:rel_acc}
\end{figure}

Next, we evaluate the relative precision by commanding the robot to mark a \SI{100}{\milli\metre} $ \times $ \SI{100}{\milli\metre} square grid of four corner markings. The resulting marked pattern is then non-deformable least-squares aligned to the nominal corner coordinates spanning from $(0,0)$ to $(100,100)$, and the deviations measured for each marked point. This mimics the precision requirements for tasks such as mounting a bracket with a square hole pattern.  

\Cref{fig:rel_acc} visualizes five repetitions of the square marking experiment. 
Overall, the average point-wise deviation from the true \SI{100}{\milli\metre} $ \times $ \SI{100}{\milli\metre} pattern is \SI{1.49}{\milli\metre}. This corresponds to the tolerance available when using M8 bolts in \SI{11}{\milli\metre} holes, a typical scenario for hole patterns in this size category (neglecting drilling tolerances).
The larger deformation of the trial marked in orange in \Cref{fig:rel_acc} can be attributed to the convergence of the state estimator between individual point markings.

\Cref{fig:acc_comparison} compares the achieved absolute accuracy and relative precision statistics. The demonstrated relative precision is close to the limits of the used hardware, as shown in our previous work (\cite{rsspaper}), allowing for consistent marking of patterns with high precision.
The larger variance and mean error of the absolute accuracy can be attributed to a stack up of state estimation and calibration inaccuracies, which can have stronger or weaker impacts depending on estimator convergence and robot orientation.

\subsection{On-site demonstration}
\begin{figure}[bt]
    \centering
    \includegraphics[width=\linewidth]{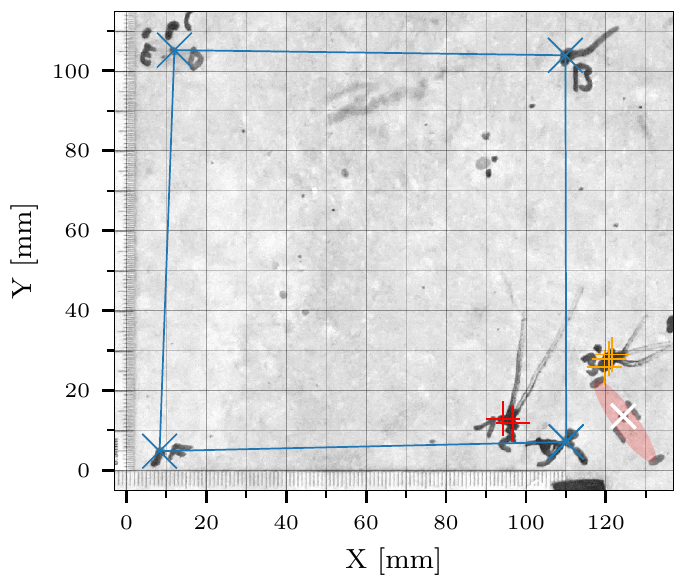}
    \caption{Qualitative illustration of on-site results under realistic operation conditions. The \SI{100}{\milli\metre} square grid is marked in blue, two different absolute precision tests (3 and 2 sequential markings) in red respectively orange. The measured target location is marked using a white cross. The red ellipse represents the laser beam divergence from the total station under the given distance and incidence angle conditions. }
    \label{fig:on_site}
\end{figure}

In addition to the laboratory environment experiments, we performed an on-site experiment on a mock construction site, as shown in \Cref{fig:full_system}. \Cref{fig:on_site} gives a qualitative overview of the results. 
The slightly worse on-site absolute precision can be attributed to an older calibration used in the on-site experiment, as it was flown before the laboratory trials.
A video of on-site trials with annotations about active policies is available in the supplementary material (Online Resource 1).

\section{Discussion}\label{sec:discussion}
The precision and accuracy are comparable to what a human worker can do on-site with modern tooling. In order to reach the quality shown here, care has to be taken during the calibration and tuning of the system. Whenever possible, we performed least-squares fitting from datasets,e.g., for intrinsics and extrinsics of cameras. More problematic are calibrations that need to be taken from CAD, such as prism-to-imu or tool frames. As an example, the tolerance of the prism mounting axis to the optical center is given as \SI{1.5}{\milli\metre}, and the exact center of percussion of the used ADIS16448B IMU is determined only to within a few millimeters (as each axis has a slightly different origin) - all these slight tolerances compound quickly. Furthermore, the state-of-the-art total station used exhibits typical beam divergence (roughly \SI{0.4}{\milli\radian}), further degrading measurements of arbitrary points at a distance.

However, one naturally wonders how precision and accuracy could be improved even more and what efforts and benefits this would entail?
After all, millimeters are tiny at the construction site scale - even the seasonal temperature fluctuation alone can cause a \SI{10}{\metre} concrete slab to contract/expand by up to \SI{4}{\milli\metre}, while typical building codes specify as-built tolerances of $5$ to \SI{10}{\milli\metre} to be within limits \cite[Norm 414]{sia414}.
A simple way to increase absolute accuracy would be to change the mechanical design such that the prism can be mounted closer and ideally rigidly to the end-effector, which is not easily possible due to visibility constraints. Likely the biggest improvement to both precision and accuracy is increasing the global yaw-estimate quality - but this could entail additional means such as fiducials, pre-scanning the construction site for SLAM, or similar, which is not always feasible.
\section{Conclusion}\label{sec:conclusion}
We presented one of the first robust and precise aerial layouting systems capable of operating under realistic conditions. In addition to our previous work, we contribute novel state estimation and navigation approaches that decouple local and global estimation and motion policies. The chosen approach has been resilient over hours of flight time, even when individual sensors failed. Through a comprehensive high-precision evaluation, we showed that the system marks at a very high relative precision of $\SI{1.5}{\milli\metre}$ and an absolute accuracy of $\SI{5.5}{\milli\metre}$. The presented approach is suitable for highly precise and robust applications on real construction sites.

\backmatter

\bibliography{sn-bibliography}
\end{document}